\documentclass[10pt,twocolumn,letterpaper]{article}

\usepackage[pagenumbers]{wacv}
\usepackage{graphicx}
\usepackage{xspace}
\usepackage{booktabs}
\usepackage{multirow}
\usepackage{array}
\usepackage{adjustbox}
\usepackage{mathtools}
\usepackage{bm}
\usepackage{microtype}
\usepackage{algorithm}
\usepackage{algpseudocode}
\usepackage{amsmath}
\usepackage{amssymb}
\usepackage{placeins}

\newcommand{\method}{TRACE\xspace}

\newcommand{\parahead}[1]{\par\noindent\textbf{#1}\ }

\definecolor{wacvblue}{rgb}{0.21,0.49,0.74}
\usepackage[pagebackref,breaklinks,colorlinks,allcolors=wacvblue]{hyperref}

\title{TRACE: Topology-Aware Reconnection via Prediction-Adaptive Connectivity Enforcement for Thin-Structure Segmentation}

\author{
\href{https://orcid.org/0009-0004-3522-6357}{\textcolor{black}{Junho Moon}}\\
Hanyang University\\
Seoul, Republic of Korea\\
{\tt\small jhmoon6807@hanyang.ac.kr}
\and
\href{https://orcid.org/0000-0002-6901-983X}{\textcolor{black}{Ikbeom Jang}}\thanks{Corresponding author.}\\
Hankuk University of Foreign Studies\\
Yongin, Republic of Korea\\
{\tt\small ijang@hufs.ac.kr}}
\date{}
\hypersetup{pdftitle={\thetitle},pdfauthor={Junho Moon, Ikbeom Jang}}

\begin{document}
\raggedbottom
\maketitle
\begin{abstract}
Thin, elongated structures are particularly sensitive to connectivity errors, as even a few local segmentation mistakes can disrupt otherwise continuous paths. Existing skeleton-based objectives typically supervise fixed centerline locations and do not account for alternative local routes that may restore connectivity. We propose \method, a topology-aware, prediction-adaptive objective that uses a target skeleton to define required local connections. For each connection, \method identifies the easiest valid reconnection under the current prediction and jointly supervises its bottleneck with the corresponding skeleton point. The supervision thus adapts to the evolving prediction rather than remaining fixed to the centerline. We further introduce a topology-aware skeleton carrier that preserves both foreground and class-specific topology, enabling reliable definition of these connections. Extensive experiments across six 2D and 3D benchmarks show that \method improves connectivity and topology while maintaining strong segmentation accuracy with low training overhead. Code will be released upon acceptance.
\end{abstract}

\section{Introduction}
\label{sec:intro}

\begin{figure}[t]
    \centering
    \includegraphics[width=1.0\linewidth]{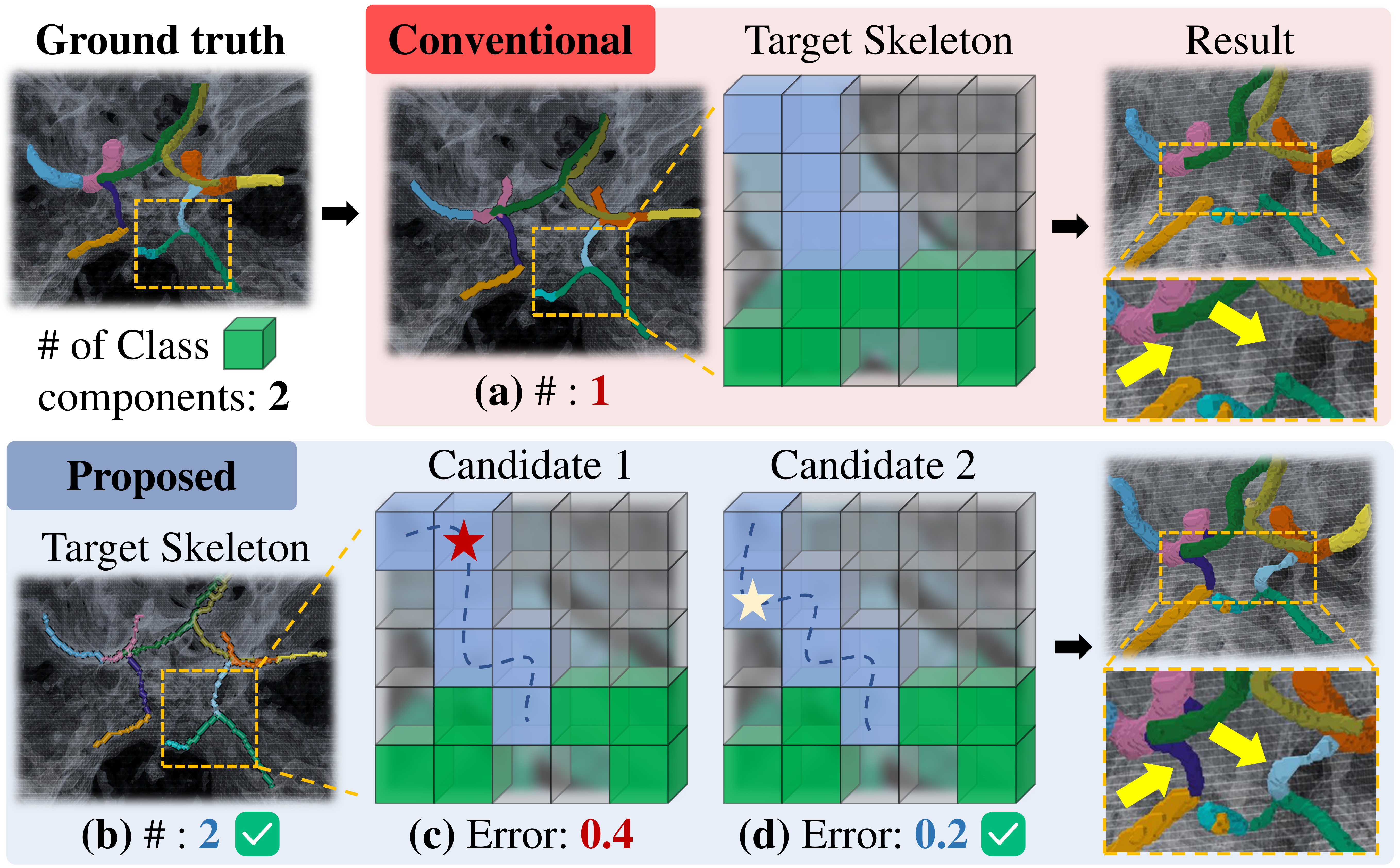}
    \caption{
    \textbf{Conventional} target-skeleton supervision \cite{kirchhoff2024skeleton} relies on fixed skeleton locations and can alter the ground-truth topology (a). \protect\\ \textbf{\method} preserves it (b), selects the easiest valid reconnection among candidates (c, d), and supervises its bottleneck (d).
    }
    \label{fig:summary}
\end{figure}

Segmentation of thin, elongated structures is particularly vulnerable to connectivity errors, as even a few misclassified pixels or voxels can break an otherwise continuous structure. This issue arises in retinal vessels \cite{staal2004ridge}, cerebral vessels \cite{xie2025dsca,yang2025benchmarking}, hepatic vessels \cite{antonelli2022msd}, pavement cracks \cite{liu2019deepcrack}, and road networks \cite{demir2018deepglobe}, where continuity is essential to structural integrity. Such local errors can fragment a single structure into multiple components while producing only minor changes in conventional region-overlap metrics \cite{shit2021cldice}. Reliable thin-structure segmentation therefore requires preserving connectivity in addition to accurate region and boundary prediction.

Prior work preserves connectivity through centerline or skeleton supervision, global topological constraints, and local relational objectives. Centerline- and skeleton-based losses \cite{shit2021cldice,acebes2024clce,shi2024cbdice} provide direct structural supervision and may additionally incorporate boundary or width information. Target-only methods \cite{kirchhoff2024skeleton} avoid skeletonizing the prediction, reducing training-time computation and memory overhead. Their supervision, however, remains confined to fixed support derived from the ground truth. More broadly, skeleton-based support depends on the chosen skeletonization procedure \cite{menten2023skeletonization}, and even topology-preserving support may not identify the prediction error responsible for a connectivity break. Global objectives based on persistent homology or component graphs \cite{hu2019topology,stucki2023betti,berger2024multiclass,lux2025topograph} constrain connected components, loops, and cavities, yet provide limited guidance for repairing a specific local break. Neighbor-, affinity-, and path-based objectives \cite{turaga2009malis,valverde2026scnp,esmaeilzadeh2025cape,zong2026wprf} provide localized supervision through individual neighbors, selected point pairs, or graph paths. These formulations organize connectivity around individual relations or pairwise paths, leaving unresolved how to jointly reconnect all same-class branches affected by a local break under the current prediction.

We propose \method, a topology-aware objective that extends skeleton supervision from a fixed target to guidance for adaptive reconnection. \method uses a topology-preserving skeleton to identify neighboring same-class branches that should remain connected. Based on the current prediction, it identifies the easiest valid local reconnection and jointly supervises the skeleton point with the bottleneck limiting this reconnection. To support this formulation, we further introduce a Topology-Aware Skeleton Carrier (TASC), which preserves both foreground and class-specific topology while defining local connectivity requirements. We identify the reconnection bottleneck through an exact local search using centered logit-margin scores derived directly from standard semantic logits. The proposed objective avoids costly prediction-side skeletonization and requires no architectural modifications, enabling efficient plug-and-play integration into existing segmentation pipelines.

\noindent Our contributions are as follows:
\begin{itemize}
\item We introduce \method, a prediction-adaptive connectivity objective that uses a skeleton to define required connections and supervises the easiest reconnection bottleneck.

\item We present TASC, a class-labeled skeleton that preserves both foreground and class-specific topology while identifying same-class branches that should remain connected.

\item Across six 2D and 3D thin-structure benchmarks, \method improves connectivity and topology while maintaining strong region-overlap and boundary accuracy.

\item We further demonstrate that \method is computationally efficient, generalizes across segmentation backbones, and complements existing topology-aware methods.
\end{itemize}

\section{Related Work}
\label{sec:related}

\parahead{Centerline and skeleton supervision.}
Centerline- and skeleton-based objectives promote connectivity by supervising thin structural representations derived from the prediction, target, or both. clDice \cite{shit2021cldice} measures reciprocal overlap between the prediction and target using differentiable soft skeletons. Menten et al. \cite{menten2023skeletonization} developed topology-preserving differentiable skeletonization for gradient-based losses. clCE \cite{acebes2024clce} applies cross-entropy at centerline locations to improve topology while preserving region accuracy. cbDice \cite{shi2024cbdice} incorporates boundary and radius cues to preserve geometry across varying widths. Skeleton Recall Loss (SkelRecall) \cite{kirchhoff2024skeleton} uses a dilated, class-labeled target skeleton and avoids prediction-side skeletonization. These methods, however, tie structural supervision to centerline or skeleton support and do not explicitly identify the prediction error that blocks a valid reconnection. \method addresses this gap with a topology-preserving carrier that defines the local same-class branches to be connected and extends pointwise supervision to the bottleneck of their easiest valid reconnection.

\parahead{Global topology supervision.}
Global topology objectives constrain connected components, loops, and cavities across the segmentation. TopoLoss \cite{hu2019topology} uses persistent homology to identify predictions associated with topological errors. Betti Matching \cite{stucki2023betti} matches topological features between the prediction and target while accounting for their spatial correspondence. Subsequent extensions support multiclass segmentation \cite{berger2024multiclass}. Topograph \cite{lux2025topograph} represents prediction--target component relations as a graph to provide efficient binary and multiclass constraints. Whereas these methods constrain global or component-level topology, \method targets a required local reconnection and directly supervises its prediction-dependent bottleneck.

\parahead{Pairwise and path supervision.}
Pairwise- and path-based objectives supervise connectivity through neighboring elements, point pairs, or graph paths. MALIS \cite{turaga2009malis} assigns pairwise connectivity errors to the weakest edge along a maximin path in an affinity graph. CAPE \cite{esmaeilzadeh2025cape} penalizes shortest-path costs between selected target-graph nodes on a predicted distance map. SCNP \cite{valverde2026scnp} adjusts semantic logits according to the weakest same-class neighbor. WPRF \cite{zong2026wprf} optimizes widest-path reachability between sampled point pairs on a target-derived support graph using a separate affinity prediction. In contrast, \method defines each local reconnection on a class-labeled carrier that preserves the topology of both the foreground union and each semantic class. Removing a carrier point exposes all neighboring same-class branches, which we jointly reconnect via an exact search over at most 9 candidate points in 2D or 27 in 3D, using a centered logit margin computed directly from standard multiclass semantic logits.

\section{Method}
\label{sec:method}

\method turns target-skeleton supervision from fixed-point enforcement into prediction-adaptive local reconnection by identifying and supervising the bottleneck of the easiest valid reconnection under the current prediction. We first formulate local reconnection as a connectivity bottleneck problem that requires a connected set of target-class points to reach all neighboring branches (\cref{sec:reconnection}). We then construct TASC, a thin class-labeled carrier that encodes these connectivity requirements while preserving both foreground and class-specific topology (\cref{sec:tasc}). Next, we define a centered logit-margin score that ranks candidate points by their rival-to-target margin relative to the carrier average (\cref{sec:class_error}). Finally, we combine the pointwise and connectivity penalties into a structural objective (\cref{sec:joint_loss}).

\subsection{Local Connectivity Bottleneck Supervision}
\label{sec:reconnection}

\begin{figure}[t]
    \centering
    \includegraphics[width=1.0\linewidth]{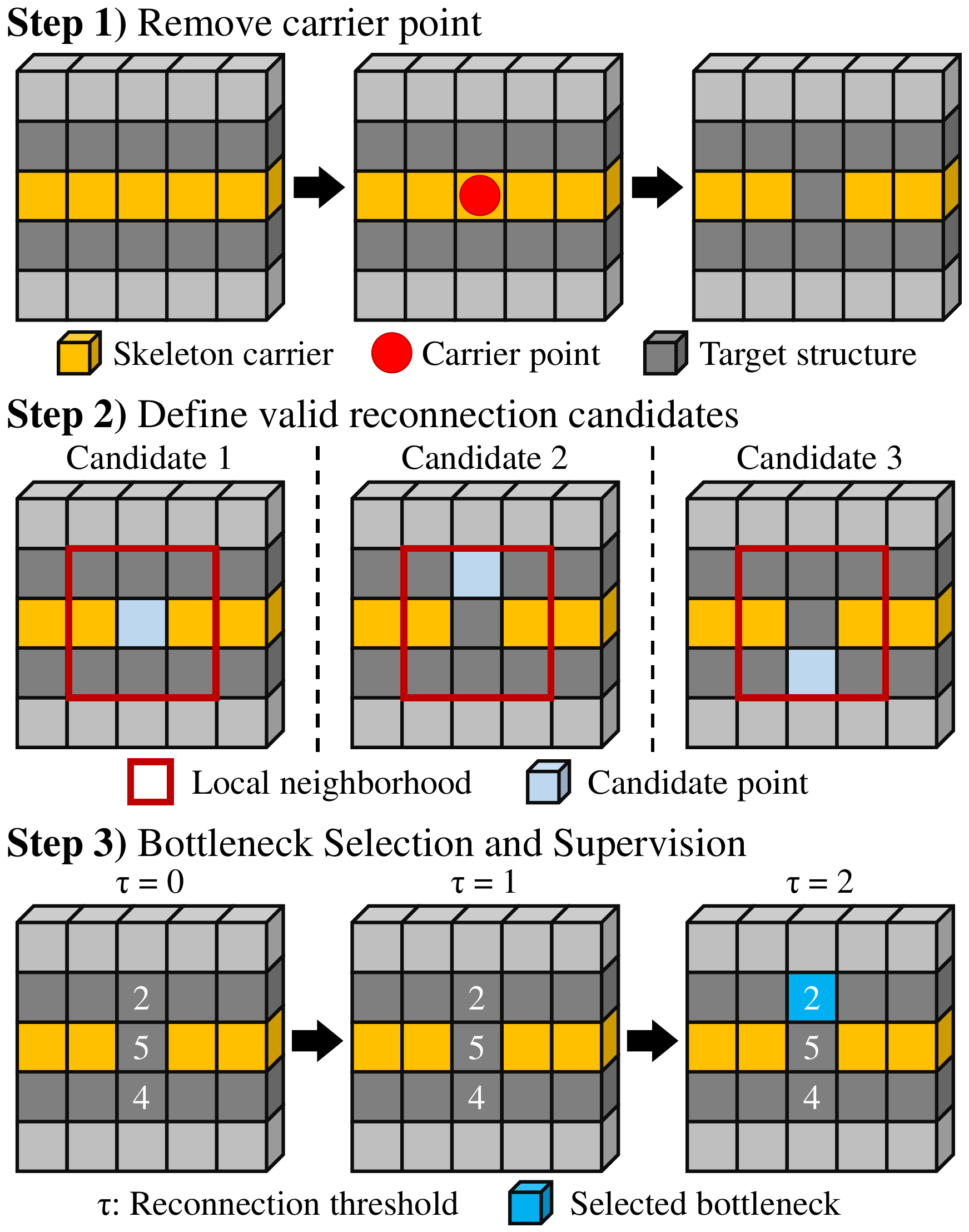}
    \caption{Prediction-adaptive local reconnection in \method. \textbf{Step 1)} Removing a carrier point exposes neighboring same-class branches that should remain connected. \textbf{Step 2)} Valid local reconnections reaching all exposed branches are considered. \textbf{Step 3)} Candidates are activated in ascending order of centered logit-margin score until a valid reconnection forms; the point activated at this threshold is the supervised bottleneck. The carrier is TASC (\cref{sec:tasc}), numbers indicate $g_c(x)$ (\cref{sec:class_error}), and $\tau$ denotes the activation threshold.}
    \label{fig:method_connectivity}
\end{figure}

We first formalize the local reconnection event supervised by \method, as illustrated in \Cref{fig:method_connectivity}. Throughout, a \emph{point} denotes a pixel or voxel, and all quantities are defined on a single output grid. Let $d\in\{2,3\}$, let $\Omega\subset\mathbb{Z}^d$ be the grid, and let $C\geq2$ be the number of semantic classes, including background. The target labeling is $Y:\Omega\to\{0,\ldots,C-1\}$.

We represent the target connectivity to be preserved using a class-labeled carrier $S:\Omega\to\{0,\ldots,C-1\}$ derived from $Y$; its construction is described in \cref{sec:tasc}. The carrier is a thin representation that preserves both foreground and class-specific topology, and its local branches define the connectivity requirements that \method supervises. For each foreground class $c$, define
\begin{equation}
K_c = \{x\in\Omega \mid S(x)=c\},
\quad
K = \bigcup_{c=1}^{C-1} K_c,
\label{eq:carrier_sets}
\end{equation}
where $K_c$ denotes the class-$c$ carrier and $K$ is the union of all foreground carriers. Removing a carrier point $v\in K_c$ reveals the neighboring same-class branches that should remain connected. For each $x\in\Omega$ with $Y(x)=c$, let $g_c(x)$ denote the centered logit-margin score defined in \cref{sec:class_error}, which ranks each target point by its rival-to-target margin relative to the carrier average.

Let $\sim_d$ denote 8-connectivity for $d=2$ and 26-connectivity for $d=3$. For a carrier point $v\in K_c$, define its local $3^d$ neighborhood as
\begin{equation}
\mathcal{N}(v) = \{x\in\Omega \mid \lVert x-v\rVert_\infty\leq1\}.
\end{equation}
Let $P_{v1},\ldots,P_{vm_v}$ be the connected components of $(K_c\cap\mathcal{N}(v))\setminus\{v\}$ under $\sim_d$; these are the \emph{neighboring same-class branches}. When $m_v\geq2$, these branches define a local reconnection event: the target requires a connected set of class-$c$ points within $\mathcal{N}(v)$ that reaches every branch. Carrier points with $m_v<2$ do not define a reconnection event.

Define the local class-$c$ target support $T_v=\{x\in\mathcal{N}(v)\mid Y(x)=c\}$ and the family of valid local reconnections
\begin{equation}
\begin{aligned}
\mathcal{R}_v = \{R\subseteq T_v\mid{}&
R\text{ is connected under }\sim_d,\\[-1pt]
& R\cap P_{vk}\neq\varnothing,\quad
\forall\,k\in\{1,\ldots,m_v\}\}.
\end{aligned}
\end{equation}
The local connectivity bottleneck is then
\begin{equation}
r_v = \min_{R\in\mathcal{R}_v}\ \max_{x\in R} g_c(x).
\label{eq:reconnection}
\end{equation}
For $m_v=2$, each $R\in\mathcal{R}_v$ contains a same-class path connecting the two branches; at a junction, one connected reconnection reaches all branches simultaneously. Thus, $r_v$ is the lowest achievable bottleneck score among valid local reconnections. A large $r_v$ means that every valid reconnection contains a point with a large rival-to-target margin relative to the carrier average.

\cref{eq:reconnection} has an equivalent threshold-connectivity form that enables exact bottleneck selection. For $\tau\geq0$, let $T_v(\tau)=\{x\in T_v\mid g_c(x)\leq\tau\}$ and let $\mathcal{C}_v(\tau)$ denote the connected components of $T_v(\tau)$ under $\sim_d$. Then
\begin{equation}
\begin{aligned}
r_v=\min\bigl\{\tau\geq0:{}&
\exists Q\in\mathcal{C}_v(\tau),\\[-1pt]
& Q\cap P_{vk}\neq\varnothing,\quad
\forall\,k\in\{1,\ldots,m_v\}\bigr\}.
\end{aligned}
\label{eq:threshold_connectivity}
\end{equation}
This characterization in \cref{eq:threshold_connectivity} yields an exact finite solver. 
We activate at most $3^d$ points in $T_v$ in nondecreasing order of $g_c$, with ties resolved by a fixed local order, and merge adjacent active components until a component reaches every branch. If $x_v^\star$ is the last activated point, then $r_v=g_c(x_v^\star)$. We perform the combinatorial search using detached scores to select $x_v^\star$, while propagating gradients through the corresponding non-detached score $g_c(x_v^\star)$. Because the carrier means in \cref{eq:class_error} remain differentiable, this update includes a direct contribution at $x_v^\star$ and the associated mean-centering correction over $K$.

\subsection{Topology-Aware Skeleton Carrier}
\label{sec:tasc}

TASC constructs a thin, class-labeled carrier whose local branches encode same-class connectivity requirements while preserving the digital topology of both the foreground union and each semantic class. Starting from $S=Y$, consider a foreground candidate point $v$ in the current carrier, with class $c=S(v)$, and define
\begin{equation}
\begin{aligned}
U(S)&=\{x\in\Omega\mid S(x)>0\},\\
U_c(S)&=\{x\in\Omega\mid S(x)=c\}.
\end{aligned}
\end{equation}
We use two standard digital-topology predicates: simple-point deletion for topology-preserving thinning~\cite{couprie2008new} and 1-isthmus detection for retaining curve-like structures exposed during thinning~\cite{bertrand2008two,couprie2016asymmetric}.

TASC preserves topology by protecting 1-isthmuses and deleting a candidate only when it is simple in both the foreground union and its class-specific support. Let $\mathsf{Simple}(v;A)$ indicate that deleting $v$ is simple in the binary set $A$, and let $\mathsf{Ist}_1(v;A)$ indicate that $v$ is a 1-isthmus of $A$. A candidate is permanently protected if it is a 1-isthmus in either $U(S)$ or $U_c(S)$; otherwise, delete it only if it is simple in both. Each accepted deletion therefore preserves the topology of the foreground union and class $c$, while leaving all other class supports unchanged. Because the predicates are evaluated on the current carrier, deletions may expose additional 1-isthmuses, which we protect when we encounter them later.

We apply these predicates using the $8/4$ foreground/background adjacency pair in 2D and $26/6$ in 3D. In 2D, the 1-isthmus predicate holds when the proper neighborhood contains exactly two 8-connected foreground components; in 3D, we use the audited DGtal $26/6$ predicate~\cite{dgtal}. \Cref{alg:tasc} summarizes the construction. The returned carrier preserves the labels of all retained points and satisfies $S(x)\in\{0,Y(x)\}$ for every $x\in\Omega$. For architectures with deep supervision~\cite{isensee2021nnunet}, we construct TASC directly from the target at each output resolution to avoid resampling-induced topological changes.

\begin{algorithm}[t]
\caption{Topology-aware skeleton carrier construction}
\label{alg:tasc}
\footnotesize
\begin{algorithmic}[1]
\Require Target $Y$, thinning order $\prec$, predicates
$\mathsf{Simple}$ and $\mathsf{Ist}_1$
\State $S\gets Y$; $\mathcal{A}\gets\varnothing$
\Repeat
    \State $\mathrm{deleted}\gets\mathrm{false}$
    \ForAll{foreground points $v\notin\mathcal{A}$ in order $\prec$}
        \State $c\gets S(v)$
        \State $U\gets\{x:S(x)>0\}$;
               $U_c\gets\{x:S(x)=c\}$
        \If{$\mathsf{Ist}_1(v;U)\lor
             \mathsf{Ist}_1(v;U_c)$}
            \State $\mathcal{A}\gets\mathcal{A}\cup\{v\}$
        \ElsIf{$\mathsf{Simple}(v;U)\land
                \mathsf{Simple}(v;U_c)$}
            \State $S(v)\gets0$
            \State $\mathrm{deleted}\gets\mathrm{true}$
        \EndIf
    \EndFor
\Until{$\mathrm{deleted}=\mathrm{false}$}
\State \Return $S$
\end{algorithmic}
\end{algorithm}

\subsection{Centered Logit Margin for Local Connectivity}
\label{sec:class_error}

The local reconnection search ranks candidate target points using a centered logit-margin score $g_c(x)$. Let $z_j(x)$ denote the logit for class $j\in\{0,\ldots,C-1\}$ at $x$. We center each logit channel by its mean over the union carrier $K$, so that the score expresses local rival-to-target margins relative to the carrier. For each class channel $j$ and each foreground target point $x$ with $Y(x)=c$, define
\begin{equation}
\begin{aligned}
\bar z_j
&= \frac{1}{|K|}\sum_{u\in K} z_j(u), \qquad
\widetilde z_j(x)=z_j(x)-\bar z_j,\\
g_c(x)
&= \left[
\max_{\ell\neq c}\widetilde z_\ell(x)
-\widetilde z_c(x)
\right]_+.
\end{aligned}
\label{eq:class_error}
\end{equation}
Here, $[a]_+=\max\{a,0\}$. Equivalently, $g_c(x)$ is the largest positive difference between a local rival-to-target margin and the corresponding carrier-average margin. This score ranks reconnection candidates, while $\mathcal{L}_{\mathrm{seg}}$ supervises the overall class prediction. In binary segmentation, the only rival is background; in multiclass segmentation, the maximization also considers other foreground classes.

\subsection{Pointwise-Connectivity Penalty Integration}
\label{sec:joint_loss}

We combine the pointwise centered logit-margin score and local connectivity bottleneck into a bounded structural penalty at each carrier point. For $a>0$, define
\begin{equation}
b(a)=2\arcsin\!\left(\tanh\frac{a}{2}\right), \quad
\phi(a)=\frac{\sqrt{2}\,a\,b(a)}{\sqrt{a^2+b(a)^2}}.
\label{eq:bounded_map}
\end{equation}
We define $\phi(0)=0$ by continuity. The mapping is increasing, has unit slope at the origin, and approaches the finite limit $\sqrt{2}\pi$. It therefore preserves sensitivity to small scores while limiting the influence of large ones.

For each carrier point $v\in K_c$, let
$p_v=\phi(g_c(v))$. When $m_v\geq2$, additionally let
$t_v=\phi(r_v)$. We define the structural penalty at $v$ as
\begin{equation}
q_v=
\begin{cases}
\displaystyle
\sqrt{\frac{p_v^2+t_v^2}{2}},
& m_v\geq2,\\[4pt]
p_v,
& m_v<2.
\end{cases}
\label{eq:joint_atom}
\end{equation}
For $m_v\geq2$, $q_v$ is the root mean square of the pointwise
and connectivity penalties, so $q_v=p_v=t_v$ whenever $p_v=t_v$.
In particular, if $x_v^\star=v$, then $r_v=g_c(v)$, and the same
local score is not double-counted.

To prevent classes with larger carriers from receiving greater
weight, we average the penalties within each nonempty class carrier
and normalize over the $C-1$ foreground classes:
\begin{equation}
\mathcal{L}_{\mathrm{TRACE}}
=
\frac{1}{C-1}
\sum_{\substack{c=1\\K_c\neq\varnothing}}^{C-1}
\frac{1}{|K_c|}
\sum_{v\in K_c} q_v.
\label{eq:reduction}
\end{equation}
The overall training objective is
\begin{equation}
\mathcal{L}
=
\mathcal{L}_{\mathrm{seg}}
+
\lambda\mathcal{L}_{\mathrm{TRACE}},
\label{eq:total}
\end{equation}
where $\mathcal{L}_{\mathrm{seg}}$ denotes a standard
segmentation loss, such as Dice loss, cross-entropy, or their
combination, and $\lambda$ controls the relative weight of
$\mathcal{L}_{\mathrm{TRACE}}$.

\section{Experiments}
\label{sec:experiments}

\subsection{Datasets and Evaluation Metrics}
\label{sec:datasets}

\parahead{Datasets.}
We evaluate \method on six thin-structure segmentation benchmarks spanning 2D and 3D data, including binary and multiclass tasks. DRIVE \cite{staal2004ridge} is used for retinal vessel segmentation and contains 40 RGB fundus images, split into 20/10/10 for training/validation/testing. DeepCrack \cite{liu2019deepcrack} contains 537 RGB images for surface-crack segmentation (270/30/237), while DeepGlobe Roads \cite{demir2018deepglobe} contains 6,226 RGB satellite images for road segmentation (4,980/622/624). For DSCA \cite{xie2025dsca}, we convert 224 cerebral digital subtraction angiography sequences, annotated for the main arterial trunk and bifurcated vessels, into 2D minimum-intensity projections and use a 144/36/44 split. MSD Hepatic Vessel \cite{antonelli2022msd} contains 303 portal-venous-phase 3D CT scans with hepatic vessel and tumor annotations; we treat tumor voxels as background for binary vessel segmentation and use a 243/30/30 split. TopCoW \cite{yang2025benchmarking} contains 125 3D CT angiography scans with multiclass annotations for 13 Circle of Willis vessel segments, split into 100/12/13 cases.

\parahead{Metrics.}
We evaluate segmentation quality using complementary metrics for connectivity, topology, boundary accuracy, and region overlap. We first report clDice \cite{shit2021cldice} to measure centerline consistency and connectivity preservation. We also report the absolute Betti number error, $\beta=|\Delta\beta_0|+|\Delta\beta_1|$ \cite{stucki2023betti}, to quantify topological errors associated with structural connectivity. The 95th-percentile Hausdorff distance (HD95) \cite{taha2015metrics} assesses boundary accuracy by measuring spatial disagreement while limiting outlier sensitivity. Finally, Dice \cite{dice1945measures} measures region overlap and assesses whether improved structural consistency is achieved without sacrificing overall segmentation accuracy. We compute each metric independently for each foreground class in every test sample. For multiclass datasets, we average the class-wise values to obtain one score per metric per sample, then average these per-sample scores over the entire test set.

\begin{table*}[t]
\centering
\caption{Comparison of training objectives on four 2D thin-structure datasets spanning binary and multiclass tasks under the same nnU-Net setup. Results are mean $\pm$ standard deviation over test samples. Bold and underlined values indicate the best and second-best results.}
\label{tab:main_2d}
\scriptsize
\setlength{\tabcolsep}{2.5pt}

\resizebox{\textwidth}{!}{\begin{tabular}{l *{2}{cccc}}
\toprule

\multirow{2}{*}{Method}
& \multicolumn{4}{c}{DRIVE~\cite{staal2004ridge}}
& \multicolumn{4}{c}{DSCA~\cite{xie2025dsca}} \\
\cmidrule(lr){2-5}
\cmidrule(lr){6-9}
& clDice$\uparrow$ & $\beta\downarrow$ & HD95$\downarrow$ & Dice$\uparrow$
& clDice$\uparrow$ & $\beta\downarrow$ & HD95$\downarrow$ & Dice$\uparrow$ \\
\midrule

Dice + CE~\cite{isensee2021nnunet}
& 81.43{\tiny$\pm$3.56}
& 115.90{\tiny$\pm$18.51}
& 4.00{\tiny$\pm$2.05}
& 82.04{\tiny$\pm$1.40}
& 87.03{\tiny$\pm$14.95}
& 62.07{\tiny$\pm$52.24}
& 49.50{\tiny$\pm$161.24}
& 86.62{\tiny$\pm$15.74} \\

+ clDice~\cite{shit2021cldice}
& 81.97{\tiny$\pm$3.69}
& 97.80{\tiny$\pm$27.52}
& 4.00{\tiny$\pm$2.19}
& \underline{82.22{\tiny$\pm$1.37}}
& 87.71{\tiny$\pm$12.69}
& \textbf{53.81{\tiny$\pm$47.27}}
& \underline{29.01{\tiny$\pm$111.52}}
& 87.25{\tiny$\pm$13.56} \\

+ Topo-clDice~\cite{menten2023skeletonization}
& \underline{81.99{\tiny$\pm$3.60}}
& \underline{82.50{\tiny$\pm$18.26}}
& \underline{3.82{\tiny$\pm$2.22}}
& 82.10{\tiny$\pm$1.41}
& \multicolumn{4}{c}{\textemdash\; Out of memory \;\textemdash} \\

+ clCE~\cite{acebes2024clce}
& 81.88{\tiny$\pm$3.45}
& 106.90{\tiny$\pm$18.81}
& 4.09{\tiny$\pm$2.33}
& 82.18{\tiny$\pm$1.31}
& 87.53{\tiny$\pm$12.51}
& 65.91{\tiny$\pm$55.25}
& 41.77{\tiny$\pm$153.70}
& 87.33{\tiny$\pm$13.36} \\

+ cbDice~\cite{shi2024cbdice}
& 81.65{\tiny$\pm$3.62}
& 109.70{\tiny$\pm$24.70}
& 4.06{\tiny$\pm$2.22}
& 82.10{\tiny$\pm$1.37}
& 87.52{\tiny$\pm$13.28}
& 61.86{\tiny$\pm$54.68}
& 41.62{\tiny$\pm$153.59}
& 87.09{\tiny$\pm$14.00} \\

+ SkelRecall~\cite{kirchhoff2024skeleton}
& 81.89{\tiny$\pm$3.50}
& 97.00{\tiny$\pm$22.78}
& 4.00{\tiny$\pm$2.35}
& 82.18{\tiny$\pm$1.27}
& \underline{87.88{\tiny$\pm$12.81}}
& 56.42{\tiny$\pm$47.61}
& 41.75{\tiny$\pm$153.69}
& \underline{87.47{\tiny$\pm$13.58}} \\

+ Betti Matching~\cite{berger2024multiclass}
& 81.60{\tiny$\pm$3.59}
& 103.70{\tiny$\pm$19.37}
& 4.30{\tiny$\pm$2.30}
& 82.11{\tiny$\pm$1.30}
& 87.42{\tiny$\pm$13.87}
& 56.63{\tiny$\pm$50.92}
& 44.80{\tiny$\pm$149.36}
& 87.12{\tiny$\pm$14.21} \\

\textbf{+ \method (ours)}
& \textbf{82.65{\tiny$\pm$3.50}}
& \textbf{79.30{\tiny$\pm$20.51}}
& \textbf{3.45{\tiny$\pm$1.77}}
& \textbf{82.38{\tiny$\pm$1.32}}
& \textbf{89.18{\tiny$\pm$10.56}}
& \underline{55.49{\tiny$\pm$50.45}}
& \textbf{25.29{\tiny$\pm$111.20}}
& \textbf{88.78{\tiny$\pm$11.06}} \\

\midrule

\multirow{2}{*}{Method}
& \multicolumn{4}{c}{DeepCrack~\cite{liu2019deepcrack}}
& \multicolumn{4}{c}{DeepGlobe Roads~\cite{demir2018deepglobe}} \\
\cmidrule(lr){2-5}
\cmidrule(lr){6-9}
& clDice$\uparrow$ & $\beta\downarrow$ & HD95$\downarrow$ & Dice$\uparrow$
& clDice$\uparrow$ & $\beta\downarrow$ & HD95$\downarrow$ & Dice$\uparrow$ \\
\midrule

Dice + CE~\cite{isensee2021nnunet}
& 89.48{\tiny$\pm$9.58}
& 5.21{\tiny$\pm$5.64}
& 11.48{\tiny$\pm$16.80}
& 82.49{\tiny$\pm$11.20}
& 81.09{\tiny$\pm$14.40}
& 11.76{\tiny$\pm$12.92}
& 139.63{\tiny$\pm$187.50}
& 74.24{\tiny$\pm$12.94} \\

+ clDice~\cite{shit2021cldice}
& 89.72{\tiny$\pm$8.92}
& 5.57{\tiny$\pm$5.96}
& \underline{11.02{\tiny$\pm$16.71}}
& 82.85{\tiny$\pm$10.09}
& \textbf{82.32{\tiny$\pm$14.26}}
& 10.27{\tiny$\pm$12.70}
& 127.55{\tiny$\pm$187.12}
& 74.69{\tiny$\pm$13.19} \\

+ Topo-clDice~\cite{menten2023skeletonization}
& \multicolumn{4}{c}{\textemdash\; Out of memory \;\textemdash}
& 81.96{\tiny$\pm$14.83}
& 9.68{\tiny$\pm$17.36}
& \underline{123.40{\tiny$\pm$197.72}}
& \underline{74.78{\tiny$\pm$14.44}} \\

+ clCE~\cite{acebes2024clce}
& 89.96{\tiny$\pm$8.59}
& 5.65{\tiny$\pm$6.35}
& 11.38{\tiny$\pm$23.92}
& 83.48{\tiny$\pm$9.78}
& 81.88{\tiny$\pm$14.39}
& 11.74{\tiny$\pm$14.41}
& 129.75{\tiny$\pm$184.46}
& 74.67{\tiny$\pm$13.00} \\

+ cbDice~\cite{shi2024cbdice}
& 89.67{\tiny$\pm$9.17}
& 6.18{\tiny$\pm$7.40}
& 13.06{\tiny$\pm$24.94}
& 82.80{\tiny$\pm$10.73}
& 81.23{\tiny$\pm$14.95}
& 11.33{\tiny$\pm$14.26}
& 128.27{\tiny$\pm$181.19}
& 73.80{\tiny$\pm$13.96} \\

+ SkelRecall~\cite{kirchhoff2024skeleton}
& 90.42{\tiny$\pm$8.59}
& \underline{5.10{\tiny$\pm$5.53}}
& 12.04{\tiny$\pm$24.95}
& 83.89{\tiny$\pm$10.21}
& 81.40{\tiny$\pm$14.80}
& 11.51{\tiny$\pm$13.19}
& 137.48{\tiny$\pm$195.13}
& 74.37{\tiny$\pm$12.73} \\

+ Betti Matching~\cite{berger2024multiclass}
& \underline{90.44{\tiny$\pm$8.56}}
& 6.11{\tiny$\pm$7.24}
& 11.62{\tiny$\pm$22.94}
& \underline{84.28{\tiny$\pm$10.22}}
& 81.69{\tiny$\pm$14.10}
& \textbf{9.10{\tiny$\pm$11.48}}
& 137.66{\tiny$\pm$189.81}
& 74.30{\tiny$\pm$12.62} \\

\textbf{+ \method (ours)}
& \textbf{90.60{\tiny$\pm$8.46}}
& \textbf{4.95{\tiny$\pm$6.19}}
& \textbf{11.02{\tiny$\pm$17.60}}
& \textbf{84.53{\tiny$\pm$9.73}}
& \underline{82.03{\tiny$\pm$14.94}}
& \underline{9.13{\tiny$\pm$12.22}}
& \textbf{117.06{\tiny$\pm$163.55}}
& \textbf{74.88{\tiny$\pm$13.22}} \\

\bottomrule
\end{tabular}}
\end{table*}

\begin{table*}[t]
\centering
\caption{Comparison of training objectives on two 3D vessel datasets spanning binary and multiclass tasks under the same nnU-Net setup. Results are mean $\pm$ standard deviation over test samples. Bold and underlined values indicate the best and second-best results.}
\label{tab:main_3d}
\scriptsize
\setlength{\tabcolsep}{2.25pt}
\resizebox{\textwidth}{!}{\begin{tabular}{l cccc cccc}
\toprule
\multirow{2}{*}{Method}
& \multicolumn{4}{c}{MSD Hepatic Vessel~\cite{antonelli2022msd}}
& \multicolumn{4}{c}{TopCoW~\cite{yang2025benchmarking}} \\
\cmidrule(lr){2-5}
\cmidrule(lr){6-9}
& clDice$\uparrow$ & $\beta\downarrow$ & HD95$\downarrow$ & Dice$\uparrow$
& clDice$\uparrow$ & $\beta\downarrow$ & HD95$\downarrow$ & Dice$\uparrow$ \\
\midrule

Dice + CE~\cite{isensee2021nnunet}
& 67.00{\tiny$\pm$16.61}
& 40.90{\tiny$\pm$24.44}
& 39.41{\tiny$\pm$34.79}
& 58.90{\tiny$\pm$15.66}
& 86.32{\tiny$\pm$9.39}
& 0.32{\tiny$\pm$0.25}
& 8.31{\tiny$\pm$6.00}
& 80.57{\tiny$\pm$7.12} \\

+ clDice~\cite{shit2021cldice}
& 68.73{\tiny$\pm$16.87}
& 38.57{\tiny$\pm$22.26}
& 28.09{\tiny$\pm$25.15}
& 59.70{\tiny$\pm$16.11}
& 85.49{\tiny$\pm$6.96}
& 0.36{\tiny$\pm$0.26}
& 7.97{\tiny$\pm$5.61}
& 81.20{\tiny$\pm$5.45} \\

+ Topo-clDice~\cite{menten2023skeletonization}
& \underline{68.92{\tiny$\pm$16.82}}
& 35.67{\tiny$\pm$20.56}
& 23.40{\tiny$\pm$27.98}
& \underline{62.83{\tiny$\pm$16.28}}
& \multicolumn{4}{c}{\textemdash\; Out of memory \;\textemdash} \\

+ clCE~\cite{acebes2024clce}
& 68.38{\tiny$\pm$16.57}
& \underline{32.13{\tiny$\pm$20.32}}
& 28.84{\tiny$\pm$28.20}
& 60.25{\tiny$\pm$15.51}
& 86.27{\tiny$\pm$7.64}
& \underline{0.29{\tiny$\pm$0.21}}
& 7.71{\tiny$\pm$6.75}
& 80.79{\tiny$\pm$6.08} \\

+ cbDice~\cite{shi2024cbdice}
& 67.61{\tiny$\pm$15.87}
& 33.77{\tiny$\pm$22.43}
& \underline{23.33{\tiny$\pm$22.54}}
& 61.70{\tiny$\pm$15.96}
& 86.19{\tiny$\pm$7.56}
& 0.30{\tiny$\pm$0.28}
& 7.30{\tiny$\pm$5.36}
& 81.20{\tiny$\pm$6.76} \\

+ SkelRecall~\cite{kirchhoff2024skeleton}
& 68.83{\tiny$\pm$17.10}
& 33.90{\tiny$\pm$20.56}
& 23.64{\tiny$\pm$32.22}
& 62.30{\tiny$\pm$16.29}
& 85.70{\tiny$\pm$9.00}
& 0.30{\tiny$\pm$0.26}
& 8.05{\tiny$\pm$7.66}
& 81.00{\tiny$\pm$7.89} \\

+ Betti Matching~\cite{berger2024multiclass}
& 66.95{\tiny$\pm$15.28}
& 33.47{\tiny$\pm$21.12}
& 24.77{\tiny$\pm$24.40}
& 61.36{\tiny$\pm$14.94}
& \underline{87.07{\tiny$\pm$8.28}}
& 0.31{\tiny$\pm$0.20}
& \underline{5.73{\tiny$\pm$7.03}}
& \underline{82.31{\tiny$\pm$6.90}} \\

\textbf{+ \method (ours)}
& \textbf{69.08{\tiny$\pm$16.75}}
& \textbf{30.77{\tiny$\pm$20.47}}
& \textbf{21.36{\tiny$\pm$34.91}}
& \textbf{63.83{\tiny$\pm$16.80}}
& \textbf{87.35{\tiny$\pm$7.75}}
& \textbf{0.26{\tiny$\pm$0.23}}
& \textbf{5.42{\tiny$\pm$5.21}}
& \textbf{82.42{\tiny$\pm$5.69}} \\

\bottomrule
\end{tabular}}
\end{table*}

\begin{figure*}[t]
    \centering
    \includegraphics[width=\textwidth]{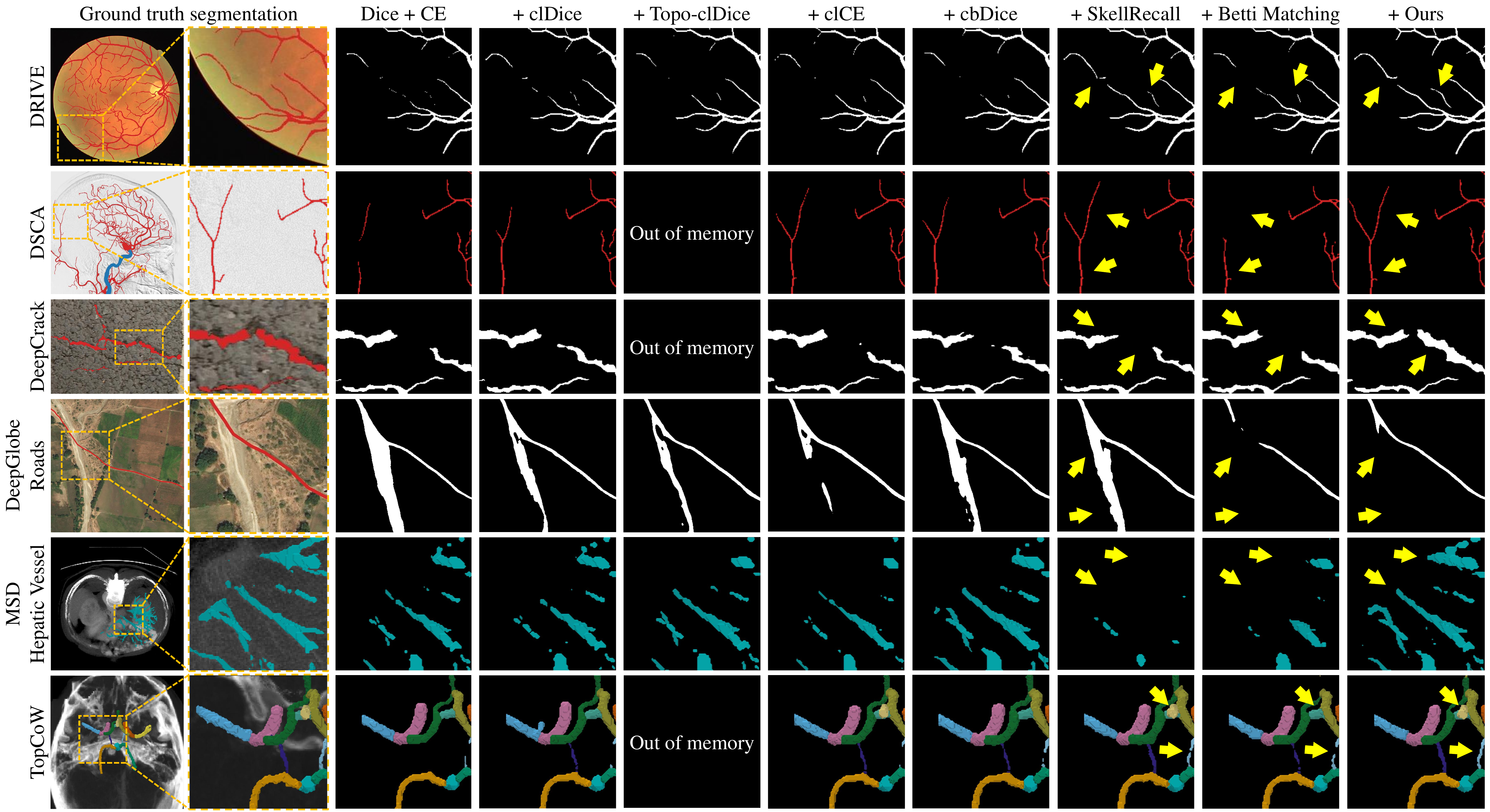}
    \caption{    
    Qualitative comparison across six 2D and 3D thin-structure datasets. Each row shows the ground truth, an enlarged region, and predictions from the compared methods. Yellow arrows highlight representative breaks, missing branches, and incorrect connections. 
    }
    \label{fig:qualitative}
\end{figure*}

\begin{figure}[t]
    \centering
    \includegraphics[width=1.0\linewidth]{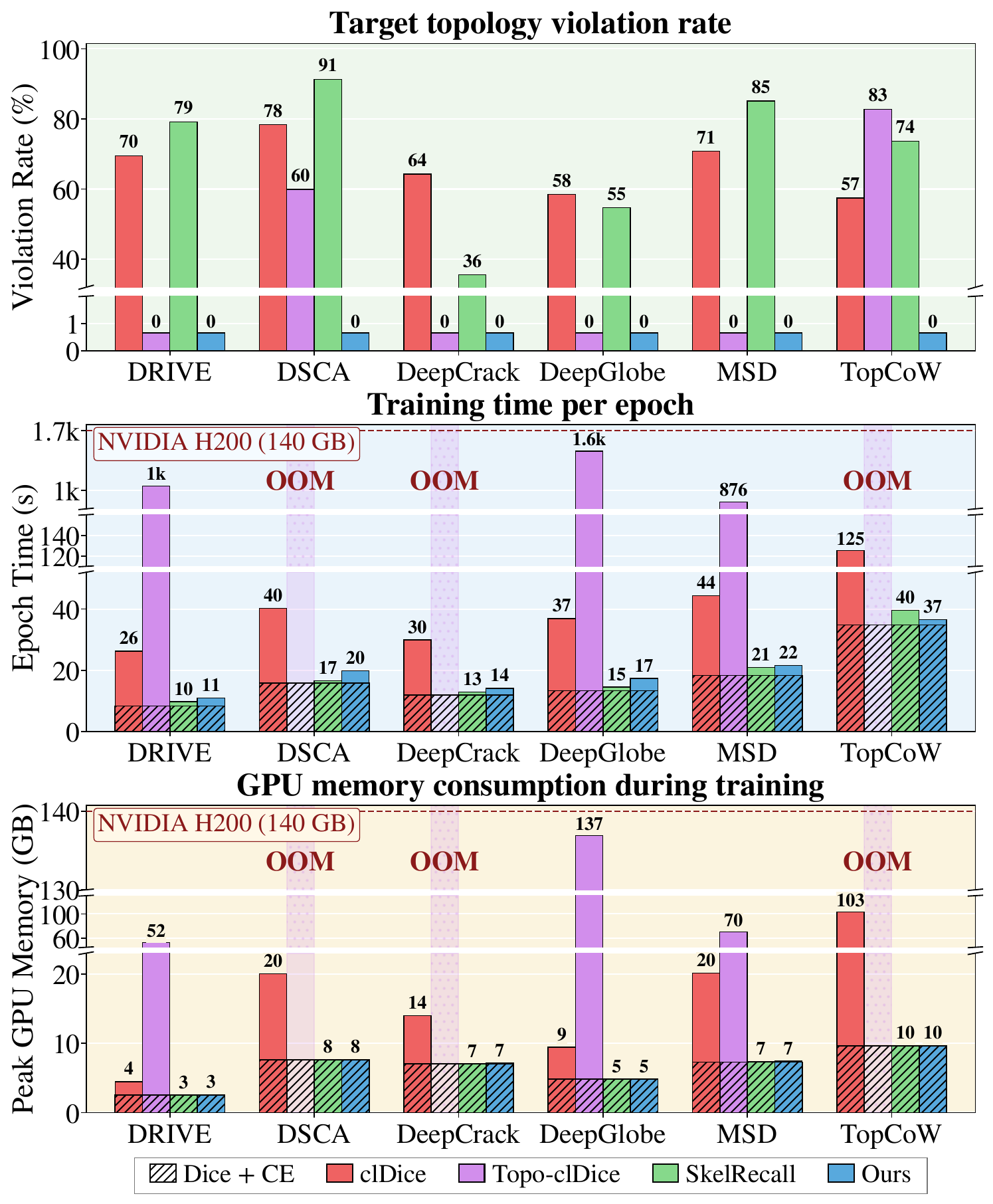}
    \caption{Topology preservation and training efficiency across six datasets. Target topology violation rate across deep-supervision scales (top); per-epoch training time and peak GPU memory on an NVIDIA H200 (middle/bottom). OOM denotes out of memory.}
    \label{fig:topo_violation}
\end{figure}

\subsection{Main Comparison Setup}
\label{sec:benchmark}

\parahead{Baselines.}
Our primary focus is connectivity preservation; therefore, we conduct extensive comparisons with representative skeleton-based losses, including clDice \cite{shit2021cldice}, clDice with topology-preserving skeletonization (Topo-clDice) \cite{menten2023skeletonization}, clCE \cite{acebes2024clce}, cbDice \cite{shi2024cbdice}, and SkelRecall \cite{kirchhoff2024skeleton}. 
We also include the recent multiclass extension of Betti Matching \cite{berger2024multiclass} as a consistent global-topology reference across binary and multiclass benchmarks, using the projection-based protocol of \cite{liu2026ag} for 3D volumes.
To isolate the effects of the training objective, the main benchmark evaluates representative methods using standard semantic outputs while keeping the prediction head and inference pipeline fixed. We exclude methods that require alternative output formulations, auxiliary or correction modules, or post-processing from this controlled comparison.

\parahead{Training protocol.}
We use a common nnU-Net \cite{isensee2021nnunet} training setup for all methods to ensure a controlled comparison. Following \cref{eq:total}, we use Dice loss combined with cross-entropy loss (Dice + CE), the default nnU-Net objective, as $\mathcal{L}_{\mathrm{seg}}$, and replace $\mathcal{L}_{\mathrm{TRACE}}$ with each compared structural loss while keeping all other training settings fixed. For \method and the skeleton-based losses, we select $\lambda$ from $\{0.10, 0.30, 0.60, 1.00\}$ based on the validation Dice score. Betti Matching is tuned separately because excessive topological weighting can degrade segmentation performance \cite{stucki2023betti}; its weight is selected on the validation set from $\{0.00001, 0.0001, 0.001, 0.01\}$.

\subsection{Topology Preservation and Training Efficiency}
\label{sec}

\parahead{Topology preservation.}
We assess whether each method's target-side skeleton representation preserves topology across deep-supervision scales. We use clDice \cite{shit2021cldice} as a representative differentiable soft-skeletonization method, Topo-clDice \cite{menten2023skeletonization} as its topology-preserving counterpart, and SkelRecall \cite{kirchhoff2024skeleton} as a target skeleton-based method. At each active scale, we construct the skeleton from the corresponding training target using each method and record a topology violation if the Betti numbers change for the foreground union or any present semantic class. We report the percentage of target--scale pairs exhibiting at least one violation.

\parahead{Training efficiency.}
We evaluate the training overhead of \method against the same representative methods under an identical nnU-Net setup, with Dice + CE as the reference. We report training time per epoch and peak process GPU memory on an NVIDIA H200 (140 GB); runs exceeding the available GPU memory are reported as out of memory (OOM). Topology preservation can still be evaluated in OOM cases, whereas training efficiency cannot be assessed without a completed epoch.

\subsection{Backbone Transfer and Method Compatibility}
We evaluate \method's plug-and-play capability in two settings: transfer to a different segmentation backbone and integration with another topology-aware method.

\parahead{Transfer to CS$^2$-Net \cite{mou2021cs2net}.}
We integrate \method with CS$^2$-Net to assess its generalization performance across segmentation architectures. We compare \method against the standard training objective and representative structural losses on DeepCrack and DRIVE.

\parahead{Integration with SCNP \cite{valverde2026scnp}.}
We further investigate whether \method provides complementary connectivity supervision when combined with an existing topology-aware approach. Specifically, we compare Dice + CE applied to the SCNP-transformed logits with the same objective augmented by \method. We conduct experiments on DRIVE and DSCA using nnU-Net.

\subsection{Ablation Study}
\label{sec:ablation}
We perform a cumulative ablation to assess the contribution of \method's main components. We compare six configurations: the Dice + CE baseline (DCE) used in nnU-Net~\cite{isensee2021nnunet}; conventional target-skeleton recall (Skel)~\cite{kirchhoff2024skeleton}; its scale-native TASC replacement (TASC); TASC with the centered logit-margin score (TASC-R); an unbounded reconnection variant with
$\phi(a)=a$ (\method-U); and the full objective incorporating exact local reconnection and bounded penalty integration (\method). We evaluate all configurations on DRIVE.

\begin{table}[t]

\centering

\caption{Backbone-transfer evaluation of \method using CS$^2$-Net on DeepCrack and DRIVE. Results are averaged over test samples; bold and underlined values indicate the best and second-best results.}

\label{tab:plugin_cs2net}

\setlength{\tabcolsep}{3.2pt}

\resizebox{\columnwidth}{!}{\par
\begin{tabular}{l l cccc}

\toprule

Dataset
& Method
& clDice$\uparrow$
& $\beta\downarrow$
& HD95$\downarrow$
& Dice$\uparrow$ \\

\midrule

\multirow{4}{*}{
\begin{tabular}[c]{@{}l@{}}
DeepCrack\\
\cite{liu2019deepcrack}
\end{tabular}}
& Dice + CE
& 89.03
& 6.16
& 16.46
& 82.78 \\

& + clDice~\cite{shit2021cldice}
& \underline{90.42}
& 5.06
& 15.75
& \underline{84.58} \\

& + SkelRecall~\cite{kirchhoff2024skeleton}
& 90.07
& \underline{4.74}
& \underline{15.68}
& 84.26 \\

& \textbf{+ \method}
& \textbf{91.10}
& \textbf{3.69}
& \textbf{15.04}
& \textbf{85.20} \\

\midrule

\multirow{4}{*}{
\begin{tabular}[c]{@{}l@{}}
DRIVE\\
\cite{staal2004ridge}
\end{tabular}}
& Dice + CE
& 82.19
& 88.80
& 4.56
& 82.27 \\

& + clDice~\cite{shit2021cldice}
& 82.58
& 73.90
& 4.04
& \textbf{82.64} \\

& + SkelRecall~\cite{kirchhoff2024skeleton}
& \underline{82.93}
& \underline{60.00}
& \underline{3.71}
& \underline{82.62} \\

& \textbf{+ \method}
& \textbf{84.07}
& \textbf{39.10}
& \textbf{3.63}
& 82.39 \\

\bottomrule

\end{tabular}\par
}

\end{table}

\begin{table}[t]
\centering
\caption{Effect of adding \method to SCNP on DRIVE and DSCA under the same nnU-Net setup. Results are averaged over test samples; bold values indicate the better result within each dataset.}
\label{tab:scnp_integration}

\resizebox{\columnwidth}{!}{\begin{tabular}{l l cccc}
\toprule
Dataset
& Method
& clDice$\uparrow$
& $\beta\downarrow$
& HD95$\downarrow$
& Dice$\uparrow$ \\
\midrule

\multirow{2}{*}{
\begin{tabular}[c]{@{}l@{}}
DRIVE\\
\cite{staal2004ridge}
\end{tabular}}
& \begin{tabular}[c]{@{}l@{}}
  Dice + CE\\
  w/ SCNP~\cite{valverde2026scnp}
  \end{tabular}
& 80.10
& 98.20
& 6.90
& 81.54 \\

& \textbf{+ \method}
& \textbf{81.48}
& \textbf{90.20}
& \textbf{4.60}
& \textbf{81.99} \\

\midrule

\multirow{2}{*}{
\begin{tabular}[c]{@{}l@{}}
DSCA\\
\cite{xie2025dsca}
\end{tabular}}
& \begin{tabular}[c]{@{}l@{}}
  Dice + CE\\
  w/ SCNP~\cite{valverde2026scnp}
  \end{tabular}
& 87.25
& 46.17
& 29.39
& 87.40 \\

& \textbf{+ \method}
& \textbf{88.44}
& \textbf{38.47}
& \textbf{27.01}
& \textbf{88.07} \\

\bottomrule
\end{tabular}}
\end{table}

\begin{table}[t]
\centering
\caption{Cumulative ablation of \method components on DRIVE. Results are averaged over test samples; bold and underlined values indicate the best and second-best results.
}
\label{tab:ablation}

\resizebox{\columnwidth}{!}{\begin{tabular}{l l cccc}
\toprule
Dataset
& Variant
& clDice$\uparrow$
& $\beta\downarrow$
& HD95$\downarrow$
& Dice$\uparrow$ \\
\midrule

\multirow{6}{*}{
\begin{tabular}[c]{@{}l@{}}
DRIVE\\
\cite{staal2004ridge}
\end{tabular}}
& DCE~\cite{isensee2021nnunet}
& 81.43
& 115.90
& 4.00
& 82.04 \\

& Skel~\cite{kirchhoff2024skeleton}
& 81.89
& 97.00
& 4.00
& 82.18 \\

& TASC
& 82.01
& 93.40
& 3.82
& 82.24 \\

& TASC-R
& 82.23
& 86.70
& 3.58
& 82.31 \\

& \method-U
& \underline{82.48}
& \underline{82.60}
& \underline{3.51}
& \underline{82.35} \\

& \textbf{\method}
& \textbf{82.65}
& \textbf{79.30}
& \textbf{3.45}
& \textbf{82.38} \\

\bottomrule
\end{tabular}}
\end{table}

\section{Result and Discussion}
\label{sec:discussion}

\parahead{Main comparison.}
\method consistently outperforms competing methods across all six benchmark datasets. As shown in \cref{tab:main_2d,tab:main_3d}, it achieves the best or near-best results in both 2D and 3D settings, covering binary and multiclass segmentation tasks. The qualitative results in \cref{fig:qualitative} further support these findings: compared with competing methods, \method produces more continuous local structures with fewer visible discontinuities while preserving the overall geometry and semantic extent of the predictions. This agreement between quantitative and qualitative results indicates that prediction-adaptive bottleneck supervision effectively improves structural connectivity. In particular, the proposed approach places structural supervision at prediction-dependent reconnection bottlenecks rather than repeatedly enforcing it at fixed skeleton locations.

\parahead{Topology preservation and training efficiency.}
\method preserves target topology across output scales while maintaining practical training efficiency. As shown in \cref{fig:topo_violation}, TASC exhibits no observed topology violations on any benchmark, whereas the compared skeleton constructions can alter the original target topology. Topo-clDice preserves the topology of individual classes; however, independently skeletonizing each class can alter the topology of the foreground union. The efficiency results further show that \method requires substantially less training time and GPU memory than differentiable soft-skeletonization alternatives, while remaining close to lightweight target-only supervision. This efficiency stems from constructing TASC solely from the target and restricting reconnection search to a small local neighborhood, making \method practical for routine use.

\parahead{Backbone transfer and method compatibility.}
\method is a plug-and-play objective that transfers across segmentation backbones and complements existing topology-aware supervision. The CS$^2$-Net results in \cref{tab:plugin_cs2net} show improved connectivity and topology over the standard objective and compared structural losses. In \cref{tab:scnp_integration}, adding \method to SCNP improves all reported metrics, demonstrating complementary benefits beyond SCNP's weakest-neighbor supervision. Because \method operates directly on standard semantic logits and target labels, it integrates into existing training pipelines without modifying the prediction head or inference procedure.

\parahead{Ablation.}
The ablation study shows that the main components of \method provide complementary improvements. As shown in \cref{tab:ablation}, replacing the conventional target skeleton with TASC improves structural consistency, while the centered logit-margin score further improves topology and boundary accuracy. Prediction-adaptive reconnection provides additional gains, and the bounded penalty integration in the full \method yields the best performance. These results suggest that topology-aware supervision, class-aware scoring, and adaptive reconnection each improve structural accuracy.

\begin{figure}[t]
    \centering
    \includegraphics[width=1.0\linewidth]{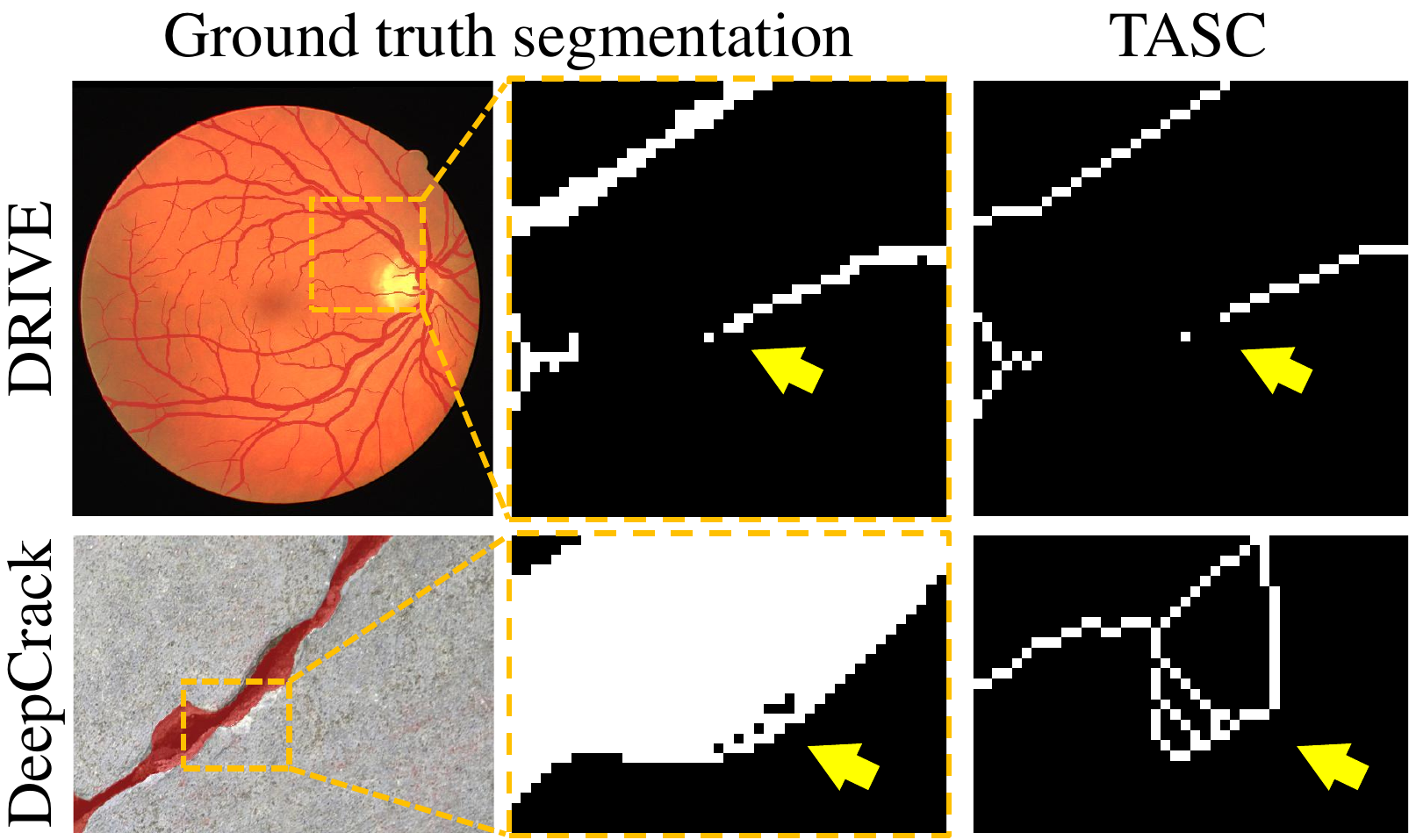}
    \caption{Annotation-induced artifacts in TASC. Small isolated regions and local irregularities in the ground-truth masks (arrows) are retained as isolated points or short loops because TASC preserves the annotated topology.}
    \label{fig:limitation}
\end{figure}

\parahead{Limitation and future work.}
\method shows consistent performance across diverse thin-structure segmentation tasks. Nevertheless, its structural supervision depends on the topology encoded in the ground-truth annotations. Small annotation errors may introduce isolated fragments, spurious branches, or local loops, which TASC preserves because removing them would alter the annotated topology, as illustrated in \cref{fig:limitation}. A promising direction is to combine TASC with uncertainty-aware filtering or consistency-based refinement to suppress unreliable topological annotations.

\section{Conclusion}
\label{sec:conclusion}

We propose \method, an efficient topology-aware objective that reframes skeleton supervision as prediction-adaptive local reconnection. TASC provides a class-labeled topological carrier that preserves foreground and class-specific topology while identifying same-class branches that should remain connected. Based on this representation, \method identifies the easiest valid reconnection under the current prediction and supervises its critical bottleneck. Across six 2D and 3D benchmarks covering binary and multiclass segmentation, \method improves connectivity and topology while maintaining region-overlap and boundary accuracy. It also preserves target topology across deep-supervision scales, introduces only modest training overhead, and requires no architectural or inference-time modifications. Its transferability across backbones and compatibility with existing topology-aware methods further demonstrate that \method is a practical plug-and-play approach for thin-structure segmentation.

\clearpage

{\small
\bibliographystyle{trace_fullname}
\bibliography{main}
}
\end{document}